\documentclass[runningheads]{llncs}
\usepackage{graphicx}
\usepackage{hyperref, color}
\usepackage[shortlabels]{enumitem}
\usepackage{subcaption}
\usepackage{multirow}
\begin{document}
\title{Digitize-PID: Automatic Digitization of Piping and Instrumentation Diagrams}
\author{Shubham Paliwal, Arushi Jain, Monika Sharma, Lovekesh Vig}
\authorrunning{S. Paliwal et al.}
\institute{TCS Research, Delhi, India \\
\email{\{shubham.p3|arushi.jain|monika.sharma1|lovekesh.vig\}@tcs.com}}
\maketitle           % typeset the header of the contribution
\vspace{-5mm}
\begin{abstract}
Digitization of scanned Piping and Instrumentation diagrams (P\&ID), widely used in manufacturing or mechanical industries such as oil and gas over several decades, has become a critical bottleneck in dynamic inventory management and creation of smart P\&IDs that are compatible with the latest CAD tools. Historically, P\&ID sheets have been manually generated at the design stage, before being scanned and stored as PDFs. Current digitization initiatives involve manual processing and are consequently very time consuming, labour intensive and error-prone. Thanks to advances in image processing, machine and deep learning techniques there is an emerging body of work on P\&ID digitization. However, existing solutions face several challenges owing to the variation in the scale, size and noise in the P\&IDs, the sheer complexity and crowdedness within the drawings, domain knowledge required to interpret the drawings  and the very minute visual differences among symbols. This motivates our current solution called \textit{Digitize-PID} which comprises of an end-to-end pipeline for detection of core components from P\&IDs like pipes, symbols and textual information, followed by their association with each other and eventually, the validation and correction of output data based on inherent domain knowledge. A novel and efficient kernel-based line detection and a two-step method for detection of complex symbols based on a fine-grained deep recognition technique is presented in the paper. In addition, we have created an annotated synthetic dataset, \textit{Dataset-P\&ID}, of $500$ P\&IDs by incorporating different types of noise and complex symbols which is made available for public use (currently there exists no public P\&ID dataset). We evaluate our proposed method on this synthetic dataset and a real-world anonymized private dataset of $12$ P\&ID sheets. Results show that Digitize-PID outperforms the existing state-of-the-art for P\&ID digitization.

%\keywords{Engineering Drawings \and Symbol Classification \and Pipeline Code Extraction.}
\end{abstract}

\section{Introduction}
\vspace{-3mm}
\label{sec:intro}
A Piping and Instrumentation Diagram (P\&ID) is a standardized schematic illustration used in the process engineering industry to record mechanical equipment, piping, instrumentation and control devices employed in the physical implementation of a process.  P\&IDs are created at the design stage of the process, stored in an image or PDF format and play an important role in the maintenance and modification stage of the physical process flow. Over the years, there are millions of PID sheets that have been manually generated, scanned and stored as images. The  valuable information trapped in these images needs to be unlocked and integrated with modern smart P\&ID systems. This digitization is necessary to facilitate easy reuse of data and design, automate mundane tasks, maintain inventory, reduce time, increase efficiency and productivity. Currently, P\&ID sheets are manually processed by engineers which is a very burdensome, time consuming and error-prone task. There is a very high cognitive load involved in manual digitization due to the minor variations in symbols, scale, size and noise within the sheets, in addition to the crowdedness of text, symbols and line. There is also significant domain knowledge involved in determining line changes and associating text with lines and symbols. Extraction and analysis of textual information, pipelines, and symbols as graphic objects and shapes are the key tasks for interpreting P\&ID sheets. We exploit the recent advances in deep learning/machine learning for these tasks.

Several approaches have been proposed for digitizing P\&ID sheets or similar documents. This includes conversion of scanned engineering drawings into 3D representation CAD files~\cite{ref_cad}, symbol recognition~\cite{ref_symrecg} and classification~\cite{ref_symclass1}, and shape representation ~\cite{ref_shaperep}. Ishii et al.~\cite{ref_ishii} presented work towards reading hand drawn piping and instrument diagram where lines, symbols and characters are separated hierarchically from the vectorized representation. 
%In another paper, Howie et al.~\cite{ref_howie} developed a vector-to-symbol recognition system to interpret the components connectivity in a CAD or paper P\&ID.
In another paper by Gellaboina et al.~\cite{ref_nn}, an iterative learning approach based on hopfield neural networks was presented to detect symbols in P\&ID sheets. 

Over the last decade, researchers have applied dynamic programming, machine learning, deep learning and pattern recognition to automate the detection of lines, text, shapes from PDFs and/or scanned images. Nazemi et al.~\cite{ref_mir} presented a method for detecting and extracting mathematical expressions, alphanumeric symbols  to generate MathML of the scanned documents. A thorough review of prior methods and a general framework for the digitization of complex engineering diagrams was proposed by Moreno-Garcia et al.~\cite{ref_moreno}. 
%They considered heuristic methods and image recognition methods for segmenting the drawing and then applied Convolution Neural Network (CNN) for classification and text interpretation. 
Fu et al.~\cite{ref_ed2em} described a visual recognition approach by leveraging CNNs for symbol recognition and methods like multi-scale sliding window and connected component analysis for automatic localization. A semi-automatic and heuristic based approach for symbol localization is proposed by Elyan et al.~\cite{ref_symclass2} which utilizes
machine learning models like Random Forests, Support Vector Machines (SVM), and CNNs. 
%The paper also highlights on applying class decomposition to identify hidden sub-classes within a symbol class.
Kang et al.~\cite{ref_kang} proposed a two-fold method comprising of extraction of relevant components from P\&IDs followed by a recognition step that compares the input sheet at various angles with the objects registered in the database. Very recently, Rahul et al.~\cite{ref_tcs} proposed a novel end-to-end approach based on a combination of low-level vision techniques and deep learning networks like CTPN~\cite{ref_ctpn} and FCN~\cite{ref_fcn} for digital interpretation of P\&ID sheets by yielding the process flow in a tree format. The shortcoming of approaches proposed in ~\cite{ref_tcs} is that it utilizes a hough transform for detecting lines which is parameter-dependent and does not perform well on noisy P\&IDs. Moreover, it uses CTPN for text detection which is not able to identify vertical text components present in P\&IDs.

\begin{figure}[h]
\centering
\includegraphics[width=0.9\textwidth]{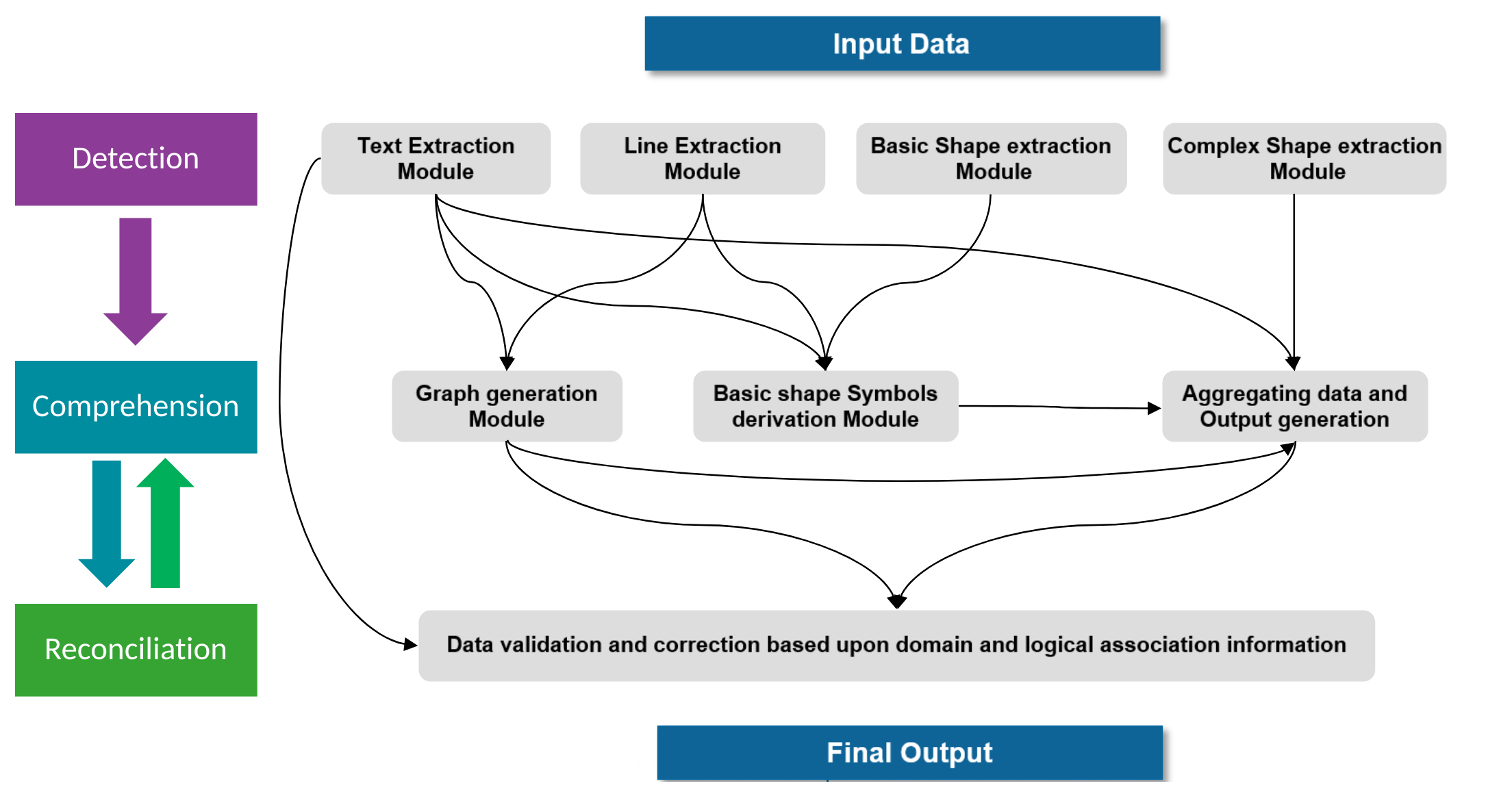}
\vspace{-3mm}
\caption{An overview of Digitize-PID which consists of $3$ sequential modules with their corresponding sub-modules: Detection, Comprehension and Reconciliation.
}
\label{fig:pipeline}
\vspace{-4mm}
\end{figure}

Although significant efforts have been made to improve the performance of automatic methods for conversion of P\&IDs into digital drawing, but perfect automatic recognition is still not achievable~\cite{ref_moreno}. To this end, we propose an end-to-end pipeline called \textit{Digitize-PID} which leverages computer-vision techniques and deep learning methods to first detect various components of interest such as lines, graphic symbols and textual information; followed by their aggregation and association with each other; and finally, validation and correction of the extracted output data based on domain rules. We describe a robust kernel-based approach for line detection which works well even in noisy environments. Additionally, a two-step process for detecting complex symbols having minute differences in visual structure is presented which utilizes a deep learning based network for symbol localization and fine-grained classification. We evaluate the effectiveness of our proposed solution on a real-world dataset of $12$ P\&ID sheets and show impressive results. Note that while $12$ may seem like a small number, each P\&ID sheet is a very high resolution image with hundreds of visual and textual components. Since, there exists no publicly available dataset for P\&ID sheets, we synthesize our own synthetic dataset named \textit{Dataset-P\&ID}\footnote{
\label{data:link}
\url{https://drive.google.com/drive/u/1/folders/1gMm_YKBZtXB3qUKUpI-LF1HE_MgzwfeR}
}. We also benchmark this dataset using Digitize-PID and make it publicly available for accelerating community advances in this field. To summarize, our key contributions in this paper are:
\vspace{-2mm}
\begin{itemize}
 \item We propose Digitize-PID, an end-to-end novel and robust pipeline for digitizing P\&ID sheets by leveraging computer vision and deep learning.
 \item Digitize-PID combines novel image-processing techniques for hard low-level vision problems such as line detection, dashed line detection, corner detection and a deep learning pipeline for symbol detection and recognition.
 \item We create a synthetic dataset of P\&ID sheets called Dataset-P\&ID consisting of $500$ P\&ID sheets with corresponding annotations for training and evaluation purposes. The dataset is released online for public use.
 \item We benchmark our proposed solution Digitize-PID on two datasets: a real-world dataset of $12$ P\&IDs and a synthetic Dataset of $100$ P\&IDs, and present the results in Section~\ref{sec:experiments}.
 \item We also compared the performance of Digitize-PID against prior state-of-the-art methods by Rahul et al.~\cite{ref_tcs} and outperformed it.
\end{itemize}

The remaining sections of the paper are structured as follows: Section~\ref{sec:methodology} describes the problem statement and discusses about the detection, comprehension, and reconciliation steps of our proposed pipeline. Section~\ref{sec:dataset} provides details about the synthetic dataset that we have generated for training and evaluation purposes. This is followed by experimental details and results in Section~\ref{sec:experiments}. Finally, we conclude the paper in Section~\ref{sec:conclusion}.

%\section{Problem Description}
%\label{sec:problemdesc}
%In this paper, the task is to automate the process of P\&ID digitization to convert the scanned legacy P\&ID sheets into database compatible format such as .csv files. The proposed method should be capable of identifying different industrial components such as symbols, pipes along with their labeled text and adjoining neighbouring symbols. For simulating the industrial scenario, we have been provided with a set of $32$ graphic symbols as shown in Figure~\ref{fig:symbol_list}. Out of them, $25$ symbols are complex symbols having very minute inter-class differences and are handled by a Complex Shape Extraction module, and remaining symbols are basic shapes with some associated regex information and can be detected using Basic Shape Extraction module. Therefore, the objective is that given an input P\&ID sheet, it should output a .csv file consisting of two separate tables - (1) listing different instances of symbols with their mapped text labels and connected pipelines; and (2) containing the list of inter-connectivity between different pipelines representing a graph.

\vspace{-3mm}
\section{Proposed Method: Digitize-PID} 
\label{sec:methodology}
\vspace{-3mm}
In this paper, the task is to automate the process of P\&ID digitization to convert the scanned legacy P\&ID sheets into a structured format. The proposed method should be capable of identifying different industrial components such as symbols, pipes along with their labeled text and  neighbouring symbols. The proposed pipeline \textit{Digitize-PID} takes an input P\&ID image and outputs a .csv file consisting of two separate tables  listing - (1)  different instances of symbols with their mapped text labels and connected pipelines; and (2) containing the list of inter-connectivity between different pipelines representing a graph.

Digitize-PID consists of three high level steps: Detection, Comprehension and Reconciliation, as shown in Figure~\ref{fig:pipeline}. The \textit{Detection} step involves extraction of different components from P\&ID sheets such as text, lines and symbols which are essential to execute the subsequent steps in the pipeline. 
%For simulating the industrial scenario, we have been provided with a set of $32$ graphic symbols as shown in Figure~\ref{fig:symbol_list}. Out of them, $25$ symbols are complex symbols having very minute inter-class differences and are handled by a Complex Shape Extraction module, and remaining symbols are basic shapes with some associated regex information and can be detected using Basic Shape Extraction module. 
The \textit{Comprehension} step consists of logically aggregating the different components detected in the previous stage, for example, the graph generation module takes basic features like lines, symbols and textual information as input and associates the appropriate symbols and text to lines. Finally, the \textit{Reconciliation} step comprises of applying different domain/business rules and final corrections/tweaks on the output data of the comprehension stage. Next, we present a detailed description of these 3 steps in the following sub-sections.

\begin{figure}
\vspace{-2mm}
\centering
\includegraphics[width=0.9\textwidth]{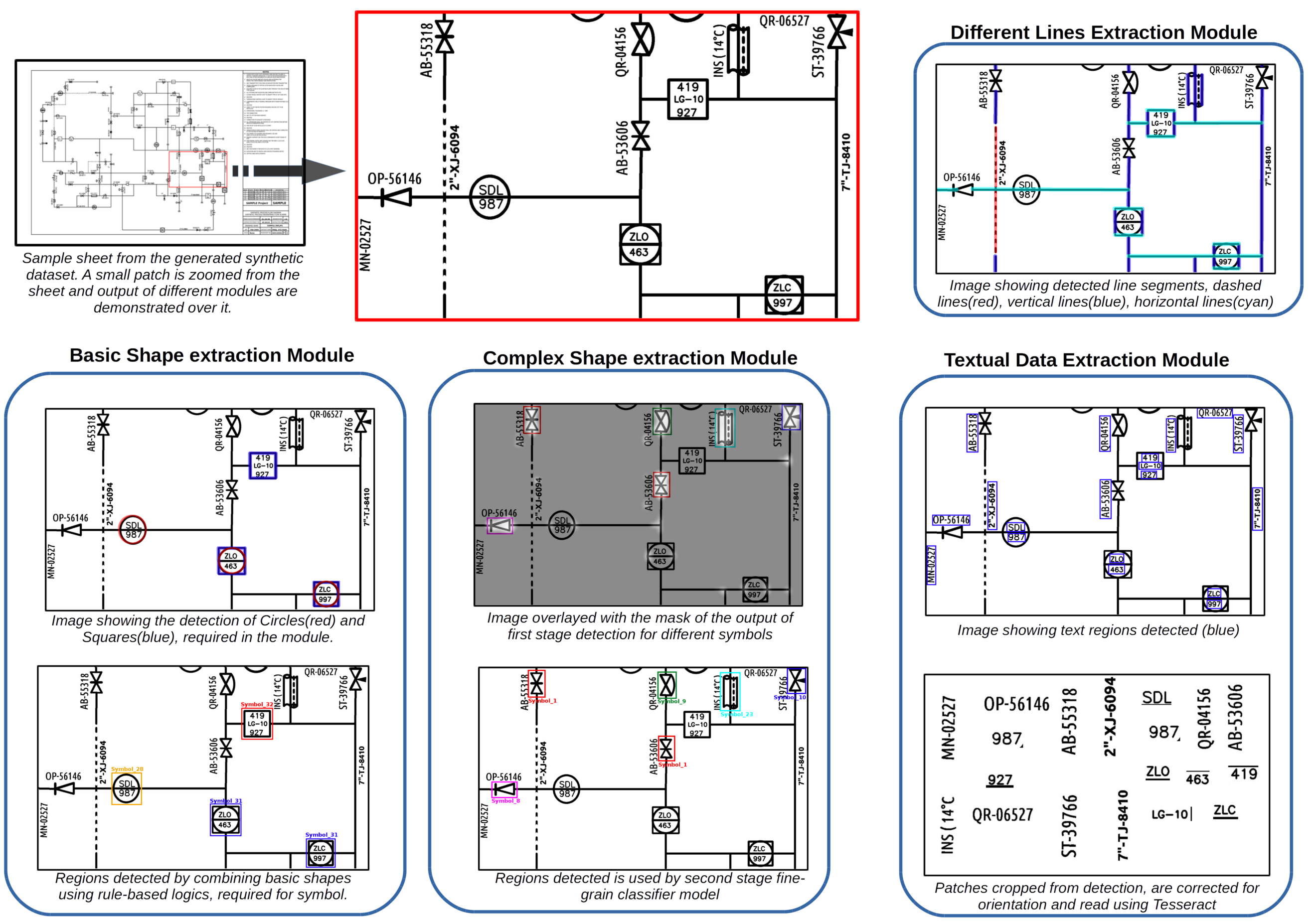}
\vspace{-2mm}
\caption{Figure illustrating different sub-modules of the Detection step of Digitize-PID over a sample P\&ID sheet (a small image-patch is zoomed for visual clarity).
}
\label{fig:line_texts_extraction}
\vspace{-2mm}
\end{figure}
\vspace{-2mm}

\subsection{Detection}
\label{subsec:detection}
The detection module comprises of the following sub-modules, as shown in Figure~\ref{fig:line_texts_extraction},
%such as text extraction, line extraction and shape extraction modules to extract various core components from P\&ID sheet, as shown in Figure~\ref{fig:line_texts_extraction}. These 
which are independent and executed in parallel.
\begin{itemize}
    \vspace{-2mm}
    \item \textbf{Text Extraction module}: P\&ID sheets contain text for labeling different components and specifying different parameters of pipelines. 
    %In addition, details of different equipment or other general instructions are given at the corner regions of the image for some cases. 
    We perform text extraction via a $2$-step process which involves dividing a P\&ID image into multiple fixed-sized overlapping patches. These patches are processed using a Character Region Awareness for Text Detection  (CRAFT)~\cite{ref_craft} network, which predicts bounding boxes (Bbox) for text regions. CRAFT works robustly even for vertically aligned texts. The overlapping Bboxes across overlapping patches are merged using IOU metric which helps to localize the text with high accuracy and effectively reduces the cases of missing texts. These merged Bboxes are projected on the input P\&ID sheet and text-patches are extracted. These patches contain single-lined texts read using Tesseract. 
    %Some outputs of text extraction module are demonstrated in Figure~\ref{fig:line_texts_extraction}.

    % \item \textbf{Line Extraction module}: A P\&ID sheet utilizes a network of different types of lines to denote connections between different components, which collectively represents the desired process flow. In our proposed Digitize-PID method, we perform line filtering using structuring elements. Since, the pixel representation of any line is a set of continuous points in a particular orientation, even a small segment of the line  is sufficient as a structuring element for finding a line of any arbitrary length at a given orientation. We choose a kernel size greater than the line width and as a function of image spatial resolution so as to avoid noise in line detection.
    \item \textbf{Line Extraction module}: A P\&ID sheet utilizes a network of different types of lines to denote connections between different components, which collectively represent the desired process flow. In Digitize-PID method, we perform line detection using filters based on a structuring element matrix. In the pixel representation, a line can be defined as set of continuous adjacent points in a particular orientation (line orientation). Thus, even an infinitesimal segment of  a line can be seen as a basic building block for the entire line. A structuring element is defined as a binary matrix of a fixed dimension ($m \times n$), in which all active regions denote the filtering line's infinitesimal segment. However, for practical purposes, we do not choose an infinitesimal segment for a structuring element matrix, rather we choose a size greater than the line width and as a function of image spatial resolution, so as to avoid noise and scaling effects in line detection.
  
    \hspace{5mm} Formally, lets assume a binary image \textit{\textbf{A}} as an integer grid \textit{\textbf{Z$^{d}$}} of dimension \textit{d} (here \textit{d}=2), and \textit{\textbf{B}} is the line structuring element belonging to the same set \textit{\textbf{Z$^{d}$}}. We first perform erosion on \textit{\textbf{A}} using \textit{\textbf{B}}, as given in Eqn.~\ref{eqn:erosion_op}. 
    %In erosion operation, \textit{\textbf{B}} is scanned over the image \textit{\textbf{A}} and union of regions having minimum value is taken as output. 
    As a result, we filter out all the elements not resembling  \textit{\textbf{B}}. Next, the filtered regions in the image are restored by performing a dilation operation, as given in Eqn.~\ref{eqn:dilation_op}, using the same structuring element \textit{\textbf{B}}. 
    %In dilation operation, \textit{\textbf{B}} is again scanned over the image \textit{\textbf{A}} and union of regions having maximum value is taken as its output. 
    \vspace{-3mm}
    \begin{equation}
    \label{eqn:erosion_op}
    \textit{\textbf{A}}^{erode} = min_{(x^{'},y^{'}):B(x^{'},y^{'})\neq0}\textit{\textbf{A}}(x+x^{'},y+y^{'})
    \end{equation}
    
    \vspace{-4mm}
    
    \begin{equation}
    \label{eqn:dilation_op}
    \textit{\textbf{A}}^{dilate} = max_{(x^{'},y^{'}):B(x^{'},y^{'})\neq0}\textit{\textbf{A}}^{erode}(x+x^{'},y+y^{'})
    \end{equation}
    
    \vspace{-6mm}

    \begin{equation}
    \label{eqn:convex_hull_op}
    conv(P) = \Big\{\sum_{i=1}^{n}\lambda_{i}p_{i}\big|\sum_{i=1}^{n}\lambda_{i}=1 \wedge \forall{i}\in{\big\{1,..,n\big\}}:\lambda_{i}\geq{0} \wedge p_{i}\in{P}\Big\}
    \end{equation}
    \vspace{-5mm}

    \hspace{5mm} Subsequently, the pixels obtained in the activated regions generate different contours over line regions. Each contour formed over set $P$, containing $n$ pixel points, is bounded by a convex hull $conv(P)$. The convex hull $conv(P)$, as defined in Eqn.~\ref{eqn:convex_hull_op}, is the intersection of all convex supersets of $P$ \cite{ref_url0}, which ensures a tight bound over the convex contour of the line. Finally, the two extreme end points from the set $conv(P)$ are computed, along the orientation of the structuring element and are treated as end points of the detected line.\\

    \vspace{-2mm}
    \textit{Dashed Line detection}: Here, we are focusing on Dashed Lines present in P\&IDs which are a series of line segments separated by equal distance, as shown in Figure~\ref{fig:line_texts_extraction}. We leverage the collinearity and consistency properties of dashed lines for  detection. There are two thresholds that we use for segment-length and distance between consecutive line-segments (gaps) which are determined based on the average value for the line cluster having the least mean segment-length and gap.
    %The proposed method uses two threshold values: one for segment length and other for in-between gaps. Since, different P\&ID sheets have different values for segment length and in-between gap, we find the average value for the cluster with least mean length/gap value for obtain generic threshold values for P\&ID sheets. 
    The cases of jumps in the series are very often noticeable in P\&ID sheets which lead to inconsistency in gaps between segments of dashed lines. This consistency is retained by applying a rule for filtering out contiguous jumps (three or more). The detected series are then merged to form the continuous chain. The only candidates for merging are the series of segments with opposite orientation and in close proximity with each other which are obtained using the DBSCAN~\cite{ref_dbscan} algorithm. 
    %Further, applying the condition on angle formed by the endpoints at the junction increases the precision of chaining.

    \item \textbf{Basic Shape extraction module}: Among various symbols used in P\&ID sheets, certain symbols are composed of primitive shapes, such as rectangles and circles (Figure~\ref{fig:symbol_list}). Some of these symbols are differentiated via the texts written inside them. One such basic shape is a circle which are detected by applying Hough transforms across different overlapping image patches followed by their aggregation.
    
   % One such very important basic shape is the circular shape, as there are many components which are built on this shape. We apply widely used Hough transform, across different overlapping patches to extract circular shapes. Finally, the circles obtained on different patches are aggregated and overlapping circles are assimilated to generate the complete list of circles. 
  
    \item \textbf{Complex Shape extraction module}: P\&ID sheets also contain very complex symbols whose structures have very minute inter-class differences and are difficult to interpret and derive via traditional image-processing. These symbols are detected using a $2$-step process which consists of a deep learning pipeline for symbol localization followed by  fine-grained recognition. As evident in Figure~\ref{fig:symbol_list}, most of the complex symbols have very similar shape, thus it is preferable to create a common class for all such symbols for symbol localization. For this, we have trained an FCN~\cite{ref_fcn} based semantic segmentation model which is used to localize all such symbols. We apply this FCN model to obtain region-proposals for symbols which are subsequently fed as input to a TBMSL-Net~\cite{ref_msml} network trained for fine-grained symbol classification. 
    %Following the two-step process of localization and recognition helps ward-off the excessive load on localization model in which it has to localise relatively fewer classes as compared to processing the complete set.
\end{itemize}

\begin{figure}
\centering
\includegraphics[width=0.8\textwidth]{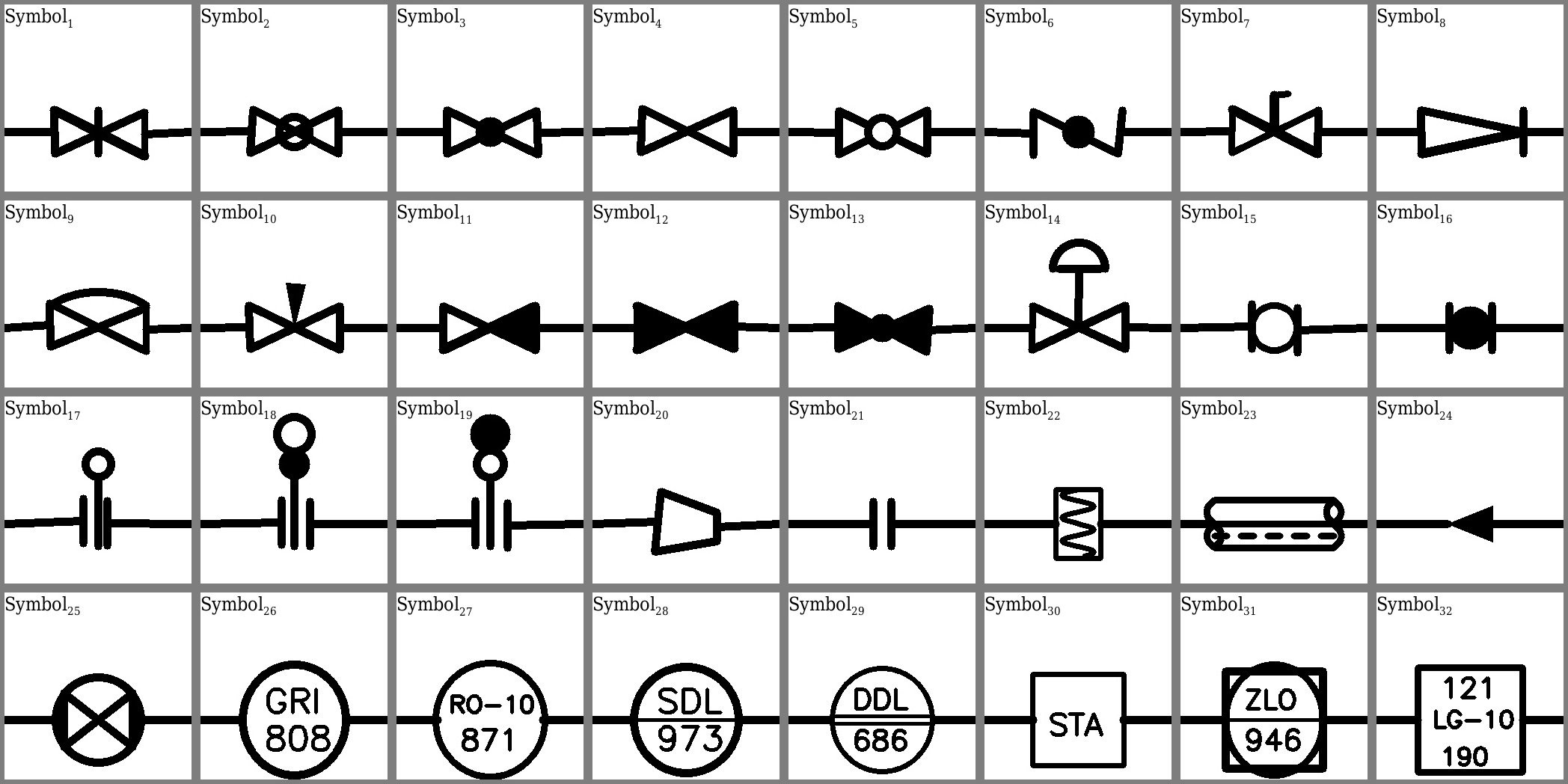}
\caption{Figure showing a set of $32$ different symbols used for Dataset-P\&ID. $Symbol_{1}$ to $Symbol_{25}$ are complex symbols as they are structurally very similar and are detected using a Complex Shape Extraction module. Remaining $Symbol_{26}$ to $Symbol_{32}$ are detected using the Basic Shape Extraction module.}
\label{fig:symbol_list}
\vspace{-2mm}
\end{figure}

\subsection{Comprehension}
    \label{subsec:comprehension}
    \vspace{-2mm}
    Now, we describe how we derive many essential properties of P\&IDs by using the appropriate logical combination of text, symbols and lines obtained previously.
    %Different components such as text, symbols and lines obtained in Section~\ref{subsec:detection} are basic features extracted from the image. However, there are many other essential properties of P\&ID which are derived by using the appropriate logical combination of one or more of these basic features. In this section, we give details about how these derived features are obtained using relevant combination of detected features.
 
    \begin{itemize}
    \item \textbf{Graph generation}: 
    %In a P\&ID, different components have their specific properties and it represents a process collectively. 
    The interactions between different components of P\&ID are represented by a web of lines that can be interpreted as a weighted graph structure. The graph representation assumes that the components are vertices and the connecting lines between components are edges. The connecting edges are of varying shapes, which can be decomposed into a combination of multiple straight lines of arbitrary lengths which enables us to create a graph with all straight line edges. For graph generation, we utilize the line information extracted in the detection step to filter out the lines with length smaller than the resolution-dependent threshold($\alpha$). Similarly, we remove the lines overlapping with regions of texts/symbols. The remaining lines (let's say $n$) are taken as edges, and the two end-points of each line are taken as vertices. This creates $n$ separate graphs, each having $2$ vertices. Thus, the junction centers having $k$ lines will have $k$ vertices, occurring in close proximity, and another $k$ vertices pointing away in different directions. These neighbouring  vertices at the junction points can be interpreted as separate clusters such that for each cluster, distance of its respective vertices from its mean (i.e., mean of vertices) would not exceed the line threshold value $\eta\alpha$ (where $0 < \eta < 1$). After optimizing cluster centers with respect to the vertices, we replace the cluster vertices with their respective cluster centers, thus aggregating the separate single-edged graphs to form a common graph.
  
    \hspace{5mm} After the graph structure is created, we assign labels to the edges. Generally, the edges labels are filtered out from the detected text using regular expressions provided by domain experts or manually by visual inspection. After the relevant labels are extracted, they are localized to corresponding graph edges which have the minimum euclidean distance. Finally, these labels on the edges are propagated using Breadth First Search, to the adjoining edges (computed for edges from left to right) with the additional stopping condition of not propagating over label-assigned edges.
  
     \item \textbf{Basic Shape Symbol detection}: 
     %As can be seen in Figure~\ref{fig:symbol_list}, some symbols are composed of basic shapes and embedded text. 
     For extracting rectangular shapes, we use the vertex sampling method to obtain candidate vertices of possible rectangular regions, which are later verified as rectangles via their geometric properties. 
     %Since, rectangles are formed using horizontal and vertical lines, intersection of horizontal and vertical lines facilitates in obtaining the vertex points. 
     The vertex points are obtained by applying the morphological \textit{AND} operator over the images of vertical and horizontal lines. Further verification of rectangle shapes is done by using the pixel values to satisfy the presence of edges across vertices. Finally, shapes which are different combinations of lines, circles and squares are logically assembled and localized. For cases where multiple symbols  have the same shape, we use the embedded-text to differentiate them. These texts circumscribed by the symbols also represent their labels.
  
     \item \textbf{Data Aggregation}: In this step, different components which include textual, graphical and symbolic information (including both complex and simple symbols) are aggregated such that each detected symbol is mapped to its label, graph nodes, and a separate identification ID is assigned. This helps to create a database of P\&ID symbols with their respective properties. Symbol mapping with graph vertices is done by using the nearest neighbour with Euclidean distance as metric. However, a similar approach cannot be used for text boxes  as they are of arbitrary length and texts found using the mean will not necessarily be closest. Thus, $k$-nearest neighbors are computed to get $k$ nearest text boxes corresponding to each symbol. Among these $k$ words, either the regex provided by domain experts are used, or else consistency in the pattern of labels is optimized over all the symbols. The symbol labels pattern which are consistent over other symbol instances in P\&ID sheet are finally assigned to the symbol. 
\end{itemize}

\vspace{-3mm}
\subsection{Reconciliation}
\vspace{-2mm}
The digitized data obtained from Comprehension step, are the final output of our proposed method. However, to address any errors/failure, we use the reconciliation step which validates and performs corrections according to domain/business rules. For example, in some arbitrary case, if the particular symbol's label has a static common name over the entire sheet, then the obtained associated text has to be re-validated and overwritten. 
%This step also facilitates user to customize the system with any required modification.
%As mentioned previously in above sections that domain experts can provide text pattern for edge labeling and symbol mapping. 
Multiple iterations involving reconciliation steps can dramatically improve the accuracy of the proposed method even in the customized business scenarios.

\begin{table}[htb]
\vspace{-5mm}
\centering
\caption{Table showing performance of Symbol Recognition on Dataset-P\&ID. \textbf{(Bottom-right)} Figure showing the confusion matrix of complex symbols detected using proposed deep learning pipeline on Dataset-P\&ID.}
\setlength\tabcolsep{0pt}
    \begin{subtable}[t]{.5\textwidth}
    	\centering
    	\vspace{-2mm}
    	\caption{\textbf{Complex Symbols}}
        \begin{tabular*}{0.9\linewidth}{@{\extracolsep{\fill}}c c c c c}
        Symbol &  Precision & Recall & F1-score \\
		\hline
        1 & 0.932 & 0.882 & 0.906\\
        2 & 0.968 & 0.968 & 0.968\\
        3 & 0.965 & 0.847 & 0.902\\
        4 & 0.974 & 0.904 & 0.938\\
        5 & 0.986 & 0.973 & 0.979\\
        6 & 0.978 & 0.967 & 0.972\\
        7 & 0.971 & 0.911 & 0.940\\
        8 & 0.823 & 0.963 & 0.888\\
        9 & 0.772 & 0.986 & 0.866\\
        10 & 0.974 & 0.958 & 0.966\\
        11 & 0.741 & 0.991 & 0.848\\
        12 & 0.875 & 0.793 & 0.832\\
        13 & 0.972 & 0.938 & 0.955\\
        14 & 0.916 & 0.961 & 0.938\\
        15 & 0.947 & 0.997 & 0.971\\
        16 & 0.979 & 0.941 & 0.960\\
        17 & 0.813 & 0.979 & 0.888\\
        18 & 0.946 & 0.993 & 0.969\\
        19 & 0.946 & 0.724 & 0.820\\
        20 & 0.962 & 0.929 & 0.945\\
        21 & 0.876 & 0.988 & 0.929\\
        22 & 0.936 & 0.946 & 0.941\\
        23 & 0.881 & 0.956 & 0.917\\
        24 & 0.977 & 0.965 & 0.971\\
        25 & 0.927 & 0.743 & 0.825\\										
        \hline
        \end{tabular*}
        \label{subtable:pr1rnn}
   \end{subtable}%
   \begin{subtable}[t]{.5\textwidth}
   		\centering
   		\vspace{-2mm}
   		 \caption{\textbf{Basic Shape Symbols}}
        \begin{tabular*}{0.9\linewidth}{@{\extracolsep{\fill}}c c c c c}
        Symbol &  Precision & Recall & F1-score \\
		\hline
        26 & 0.893 & 0.937 & 0.914\\
        27 & 0.864 & 0.903 & 0.883\\
        28 & 0.961 & 0.975 & 0.968\\
        29 & 0.977 & 0.984 & 0.980\\
        30 & 0.890 & 0.912 & 0.901\\
        31 & 0.904 & 0.892 & 0.898\\
        32 & 0.923 & 0.948 & 0.935\\
        \hline
        & & & \\   
        & & & \\             
		\multicolumn{4}{c}{\multirow{5}{*}{\includegraphics[width=2.0in,height=1.9in]{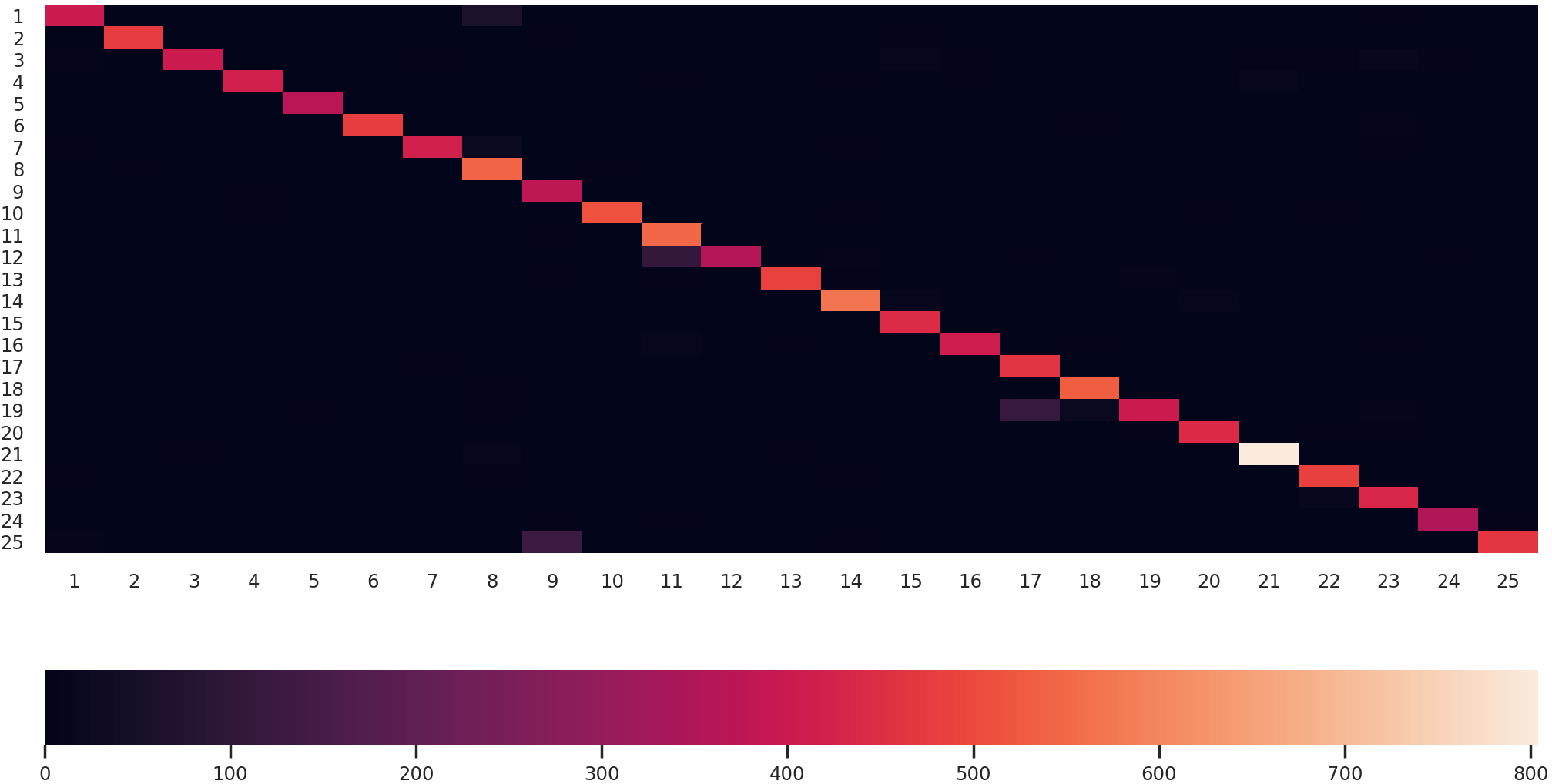}}} \\
        
        \end{tabular*}
        \label{subtable:pr2rnn}
    \end{subtable}
    \label{symbol_table}
    \vspace{-3mm}
\end{table}

\vspace{-3mm}
\section{Dataset}
\label{sec:dataset}
\vspace{-2mm}
Since, there exists no publicly available dataset for P\&ID sheets, we have generated our own synthetic dataset named \textit{Dataset-P\&ID} for training and evaluation purposes. Dataset-P\&ID consists of $500$ annotated P\&ID sheets with a 4:1 train-test ratio and is made publicly available for the benefit of research community. It includes $32$ different symbols, as given in Figure~\ref{fig:symbol_list}, uniformly plotted over different graph structures which have been generated to mimic real world P\&ID sheets as we have introduced different types of noise such as pixelation, blurring, salt and pepper noise in the generated sheets. The labels are assigned to symbols and pipelines maintaining the standards followed for real world P\&ID. The ground-truth of the dataset consists of spatial information of symbols along with associated text labels and their connected pipeline. 
%We do not provide ground truth for graph to avoid ambiguity as the graph is generally subject to different constraints. Therefore, 
We provide sets of horizontal and vertical lines with their coordinates and a separate list containing all the texts present in P\&ID sheets along with their spatial position.

\section{Experimental Results and Discussions}
\label{sec:experiments}
\vspace{-3mm}
Here, we present the system configuration used for conducting experiments followed by the performance evaluation of Digitize-P\&ID. The performance is evaluated based on Recall, Precision and F1-score for different symbols taken over the test-split. A correct prediction of a symbol includes precise localization of symbol with $IOU > 0.75$, symbol class and its associated text-label. Similarly, the output graph is evaluated based on the accuracy of correct adjacency list. However, the validation of graph-creation depends on domain information and is performed as part of the reconciliation step.

%However, graph creation depends more upon the domain information, where compatibility is verified. Therefore, we have not created any benchmark on graph metric, as it mostly involves the reconciliation step.

\begin{figure}[!h]
\centering
\includegraphics[width=0.7\textwidth]{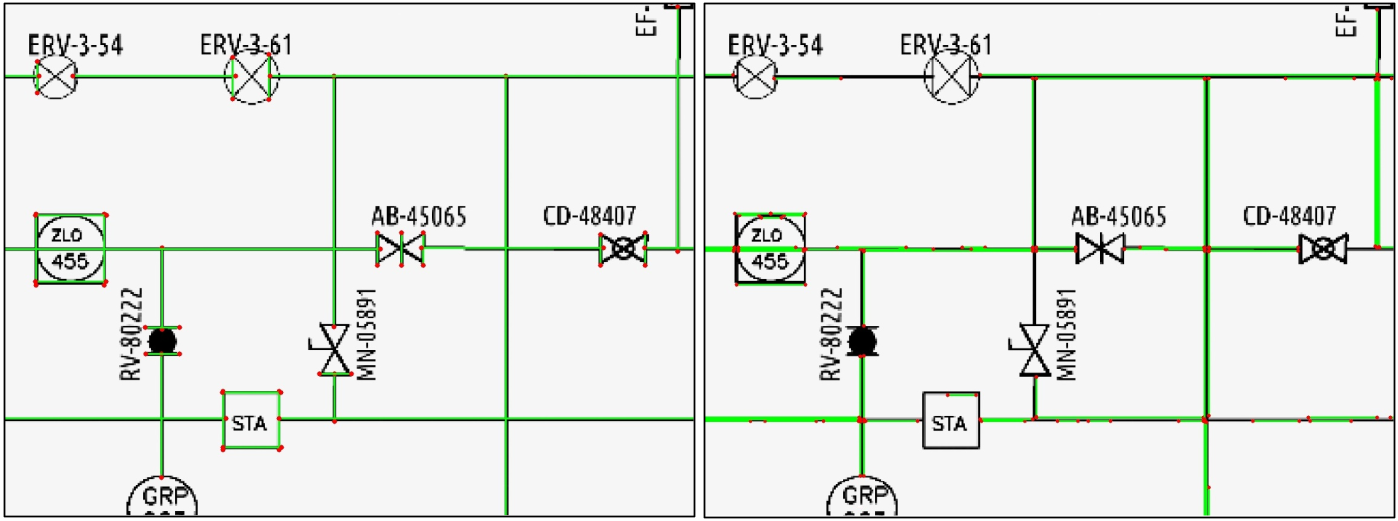}
\caption{\textbf{Left} image shows our structuring element based line-detection output and \textbf{Right} image shows hough line-detection~\cite{ref_tcs} output.}
\label{fig:line-comp}
\vspace{-2mm}
\end{figure}

\vspace{-2mm}
\begin{enumerate}[label=(\textbf{\alph*})]
\item \textbf{Setup}: We have validated and refined our proposed pipeline via repeated experiments to identify optimal parameters. To begin with, we resize the images to have width of $7168$ pixels while maintaining the aspect-ratio. In the detection module, for text detection we split the image into multiple square patches of dimension $800$ pixels such that there is an overlap of $50\%$ with their adjacent patches. The common text regions are segregated and read by using Tesseract with line configuration. The same process is also applied on the image by rotating it to capture missing vertical text. Next, we process the entire image at once for line detection. As mentioned earlier, the choice of kernel length is taken as $0.1\%$ of the maximum image resolution. Similarly, in the Basic shape extraction module, the choice of range of radius for hough circle detection is also taken between $0.05\%$ and $0.01\%$ of maximum image resolution. Further, in case of complex shape detection, the image is processed at patch level with patches of size $400$px. The output of the FCN model is filtered with threshold probability of $0.8$. Finally, the recognition threshold of TBMSL-Net is taken as $0.9$ for identifying a symbol from a region-of-interest of image. In the comprehension step, the graph is generated as explained earlier, with the DBSCAN threshold of $50$ and the neighbour threshold of $2$. The pipeline labels are spread across different pipelines using the Breadth first search algorithm. In data aggregation, we use the standard approach of connecting the nearest line entity. However for texts, we used $5$ nearest neighbours, followed by the mapping in accordance with the symbol label rule provided in the reconciliation step. In the absence of such rules, the nearest label texts following the pattern is determined as the associated texts.

\item \textbf{Results \& Discussion}: We first present the overall performance of symbol detection with correct associations on synthetic Dataset-P\&ID in Table~\ref{subtable:pr1rnn} and Table (\ref{subtable:pr2rnn}) for the complex and basic shape symbols, respectively. We also show the confusion matrix to demonstrate the robustness of our proposed complex symbol detection module on the synthetic dataset. We use the nearest associated text to resolve the conflict of multiple symbols. 

\vspace{-5mm}
\begin{table}[!h]
\centering
\caption{Comparison of Digitize-P\&ID with prior-art~\cite{ref_tcs} on $12$ real-world P\&IDs}
\label{table:tableComp}
%\begin{tabular}{|l|l|l|l|l|l|l|}
\begin{tabular}{|p{2cm}|p{1cm}|p{1cm}|p{1cm}|p{1cm}|p{1cm}|p{1cm}|}
\hline & \multicolumn{2}{c|}{\textbf{Precision}}   & \multicolumn{2}{c|}{\textbf{Recall}} & \multicolumn{2}{c|}{\textbf{F1-score}} \\ 
\hline \textbf{Symbols} & \multicolumn{1}{c|}{\textbf{\cite{ref_tcs} }} & \multicolumn{1}{c|}{\textbf{Ours}} & \multicolumn{1}{c|}{\textbf{\cite{ref_tcs} }} & \multicolumn{1}{c|}{\textbf{Ours}} & \multicolumn{1}{c|}{\textbf{\cite{ref_tcs} }} & \multicolumn{1}{c|}{\textbf{Ours}}  \\ 
\hline

\hline
\textbf{Bl-V} & 0.925 & \textbf{0.963} & 0.936 & \textbf{0.986} & 0.931 & \textbf{0.974} \\
\textbf{Ck-V} & 0.941& \textbf{0.968} & 0.969 & \textbf{0.988} & 0.955 & \textbf{0.978}\\
\textbf{Ch-sl} & \textbf{1.000} & 0.990 & 0.893 & \textbf{0.946} & 0.944 & \textbf{0.967}\\
\textbf{Cr-V} & 1.000 & 1.000 & \textbf{0.989} & 0.973 & 0.995 & \textbf{0.986}\\
\textbf{Con} & \textbf{1.000} & 0.975 & 0.905 & \textbf{0.940} & 0.950 & \textbf{0.957}\\
\textbf{F-Con} & 0.976 & 0.976 & 0.837 & \textbf{0.905} & 0.901 & \textbf{0.939}\\
\textbf{Gt-V-nc} & 0.766 & \textbf{0.864} & 1.000 & 1.000 & 0.867 & \textbf{0.927}\\
\textbf{Gb-V} & 0.888 & \textbf{0.913} & 0.941 & \textbf{0.946} & 0.914 & \textbf{0.929}\\
\textbf{Ins} & 1.000 & 1.000 & \textbf{0.985} & 0.964 & \textbf{0.992} & 0.982\\
\textbf{GB-V-nc} & 1.000 & 1.000 & 0.929 & \textbf{0.936} & 0.963 & \textbf{0.967}\\
\textbf{Others} & 0.955 & \textbf{0.973} & \textbf{1.000} & 0.990 & 0.970 & \textbf{0.981}\\
\hline

\end{tabular}
\vspace{-3mm}
\end{table}
\vspace{-4mm}

\hspace{5mm}Next, we compare our results with Rahul et.al~\cite{ref_tcs}. The symbol detection accuracy is compared on the same set of symbols, used in~\cite{ref_tcs}, on the $12$ real P\&ID sheets dataset as given in Table~\ref{table:tableComp}. It shows that the F1-score of complex symbol detection module of Digitize-PID is better than prior art~\cite{ref_tcs}. Please note that the experiment is conducted with all the symbols, but the network is only trained to identify the concerned symbols and the remaining symbols are grouped into an 'others' class. We also illustrate that our proposed structuring element based line detection is more robust than the hough line detection used in~\cite{ref_tcs}, as shown in Figure~\ref{fig:line-comp}. Finally, we present the performance of text detection and recognition on our dataset in Table~\ref{subtable:texts}. The IOU threshold value is used to find texts having significant overlap with ground truth which are used for further recognition. Since the text-labels contain very critical information, we consider a  prediction to be correct only when there is an exact match with the ground-truth label. Table~\ref{subtable:lines} shows the accuracy for line detection, for both complete ($99.34\%$) and dashed ($82.91\%$) lines (since line detection is not-learning based, results are computed over entire dataset).
 
\vspace{-7mm}
\begin{table}[!h]
\caption{Performance of Digitize-PID on synthetic Dataset-P\&ID}
\vspace{-2mm}
\setlength\tabcolsep{0pt}
    \begin{subtable}[t]{.45\textwidth}
    	\centering
    	\caption{Performance of Text Detection and Recognition module}
        \begin{tabular*}{0.9\linewidth}{@{\extracolsep{\fill}}c c c c}
        \textbf{IOU} &  \textbf{$Acc_{Det}$} & \textbf{$Acc_{Rec}$} \\
		\hline
		$<$0.9 & 87.18\% & 79.21\% \\
        \hline
        \end{tabular*}
        
        \label{subtable:texts}
   \end{subtable}%
   \hspace{5mm}
   \begin{subtable}[t]{.45\textwidth}
   		\centering
   		\caption{Performance of Dashed and Complete Line Detection module}
        \begin{tabular*}{0.9\linewidth}{@{\extracolsep{\fill}}c c c c c}
        \textbf{Type} &  \textbf{Correct}  & \textbf{Accuracy} \\
		\hline
		Complete & 90774/91416 & 99.34\% \\
		Dashed & 20620/24848 &  82.91\% \\
        \hline
        \end{tabular*}
        \label{subtable:lines}
    \end{subtable}
    \vspace{-5mm}
\end{table}
\vspace{-3mm}

\end{enumerate}

\vspace{-4mm}

\vspace{-3mm}
\section{Conclusion}
\label{sec:conclusion}
\vspace{-3mm}
In this paper, we have proposed a complete automated pipeline for extracting relevant information from P\&IDs, which are commonly used across several industry verticals for depicting a formal process flow. The proposed pipeline, named Digitize-PID utilizes a combination of state-of-the-art methods for text recognition, robust line detection using morphological operations and a two-step deep-learning based pipeline for fine-grained symbol detection and recognition. Finally, we combine all the extracted information in a graph and organize the extracted data into database-compatible tables. In addition to this, we have synthesized a dataset for P\&IDs (Dataset-P\&ID) along with their ground-truth annotations which is made public for validation by other researchers.

%Also, we have tried to address major roadblock for researchers operating in field of automating engineering drawings, as there is no standard benchmarking datasets. As per the best of our knowledge, we have made the first such attempt to create and release P\&ID dataset in public. In future, we will be releasing more realistic P\&ID dataset to further boost research in this exciting research area.
\vspace{-3mm}

\small
\bibliographystyle{splncs04}
\bibliography{mybibliography}

\begin{thebibliography}{10}
\providecommand{\url}[1]{\texttt{#1}}
\providecommand{\urlprefix}{URL }
\providecommand{\doi}[1]{https://doi.org/#1}

\bibitem{ref_symclass1}
Ablameyko, S., Uchida, S.: Recognition of engineering drawing entities: Review
  of approaches. International Journal of Image and Graphics  \textbf{7},
  709--733 (10 2007)

\bibitem{ref_craft}
Baek, Y., Lee, B., Han, D., Yun, S., Lee, H.: Character region awareness for
  text detection (craft). Conference on Computer Vision and Pattern Recognition
  (CVPR)  (2019), \url{https://arxiv.org/abs/1904.01941}

\bibitem{ref_symrecg}
Cordella, L., Vento, M.: Symbol recognition in documents: A collection of
  techniques? IJDAR  \textbf{3},  73--88 (12 2000). \doi{10.1007/s100320000036}

\bibitem{ref_symclass2}
{Elyan}, E., {Garcia}, C.M., {Jayne}, C.: Symbols classification in engineering
  drawings. In: 2018 International Joint Conference on Neural Networks (IJCNN).
  pp.~1--8 (2018). \doi{10.1109/IJCNN.2018.8489087}

\bibitem{ref_dbscan}
Ester, M., Kriegel, H.P., Sander, J., Xu, X.: A density-based algorithm for
  discovering clusters in large spatial databases with noise. In: Proceedings
  of the Second International Conference on Knowledge Discovery and Data
  Mining. p. 226–231. KDD'96, AAAI Press (1996)

\bibitem{ref_ed2em}
Fu, L., Kara, L.: From engineering diagrams to engineering models: Visual
  recognition and applications. Computer-Aided Design  \textbf{43},  278--292
  (03 2011). \doi{10.1016/j.cad.2010.12.011}

\bibitem{ref_nn}
Gellaboina, M., Venkoparao, V.: Graphic symbol recognition using auto
  associative neural network model. pp. 297--301 (02 2009).
  \doi{10.1109/ICAPR.2009.45}

\bibitem{ref_ishii}
Ishii, M., Ito, Y., Yamamoto, M., Harada, H., Members, M.: An automatic
  recognition system for piping and instrument diagrams. Systems and Computers
  in Japan  \textbf{20},  32 -- 46 (09 2007). \doi{10.1002/scj.4690200304}

\bibitem{ref_kang}
Kang, S.O., Lee, E.B., Baek, H.K.: A digitization and conversion tool for
  imaged drawings to intelligent piping and instrumentation diagrams (p\&id).
  Energies  \textbf{12}, ~2593 (07 2019). \doi{10.3390/en12132593}

\bibitem{ref_cad}
Kanungo, T., Haralick, R.M., Dori, D.: Understanding engineering drawings: A
  survey

\bibitem{ref_fcn}
Long, J., Shelhamer, E., Darrell, T.: Fully convolutional networks for semantic
  segmentation. CoRR  \textbf{abs/1411.4038} (2014),
  \url{http://arxiv.org/abs/1411.4038}

\bibitem{ref_moreno}
Moreno-garc{\'i}a, C., Elyan, E., Jayne, C.: New trends on digitisation of
  complex engineering drawings. Neural Computing and Applications
  \textbf{31}(6),  1695--1712 (Jun 2019). \doi{10.1007/s00521-018-3583-1}

\bibitem{ref_mir}
Nazemi, A., Murray, I., Mcmeekin, D.: Mathematical information retrieval (mir)
  from scanned pdf documents and mathml conversion. IPSJ Transactions on
  Computer Vision and Applications  \textbf{6},  132--142 (01 2014).
  \doi{10.2197/ipsjtcva.6.132}

\bibitem{ref_tcs}
Rahul, R., Paliwal, S., Sharma, M., Vig, L.: Automatic information extraction
  from piping and instrumentation diagrams. In: ICPRAM (2019)

\bibitem{ref_ctpn}
Tian, Z., Huang, W., He, T., He, P., Qiao, Y.: Detecting text in natural image
  with connectionist text proposal network (2016)

\bibitem{ref_url0}
Weisstein, E.W.: "convex hull.". From MathWorld--A Wolfram Web Resource.
  \url{https://mathworld.wolfram.com/ConvexHull.html}

\bibitem{ref_shaperep}
Zhang, D., Lu, G.: Lu, g.: Review of shape representation and description
  techniques. pattern recognition 37, 1-19. Pattern Recognition  \textbf{37},
  1--19 (01 2004). \doi{10.1016/j.patcog.2003.07.008}

\bibitem{ref_msml}
Zhang, F., Zhai, G., Li, M., Liu, Y.: Three-branch and mutil-scale learning for
  fine-grained image recognition (tbmsl-net). arXiv preprint arXiv:2003.09150
  (2020), \url{http://arxiv.org/abs/2003.09150}

\end{thebibliography}
\end{document}